\documentclass[12pt,reqno]{amsart}
\usepackage{latexsym}
\usepackage{amsmath}
\usepackage{braket}
\usepackage{amscd}
\usepackage{amssymb}
\usepackage{mathrsfs}
\usepackage{comment}
\usepackage{color}
\usepackage{enumerate}

\newcommand{\RR}{\mathbb{R}}
\usepackage{graphicx}
\usepackage{subfigure}
\usepackage{url}

\numberwithin{equation}{section}

\begin{document}

\title{A Stochastic Large Deformation Model for Computational Anatomy }

\author{A. Arnaudon$^{1}$, D.D. Holm$^{1}$, and A. Pai$^{2}$, S. Sommer$^{2}$}%

\address{$^{1}$ Department of Mathematics, Imperial College, London SW7 2AZ, UK.\\
$^{2}$ Department of Computer Science (DIKU), University of Copenhagen, Denmark}

\begin{abstract}
In the study of shapes of human organs using computational anatomy, variations are found to arise from inter-subject anatomical differences, disease-specific effects, and measurement noise. This paper introduces a stochastic model for incorporating random variations into the Large Deformation Diffeomorphic Metric Mapping (LDDMM) framework. By accounting for randomness in a particular setup which is crafted to fit the geometrical properties of LDDMM, we formulate the template estimation problem for landmarks with noise and give two methods for efficiently estimating the parameters of the noise fields from a prescribed data set. One method directly approximates the time evolution of the variance of each landmark by a finite set of differential equations, and the other is based on an Expectation-Maximisation algorithm. In the second method, the evaluation of the data likelihood is achieved without registering the landmarks, by applying bridge sampling using a stochastically perturbed version of the large deformation gradient flow algorithm. The method and the estimation algorithms are experimentally validated on synthetic examples and shape data of human corpora callosa.

\end{abstract}

\maketitle 
\section{Introduction}

Computational anatomy (CA) concerns the modelling and computational analysis of shapes of human organs. In CA, observed shapes or images of shapes exhibit variations due to multiple factors, such as inter-subject anatomical differences, disease-specific effects, and measurement noise. 
These variations in anatomy occur naturally over all populations and they must be handled in cross-sectional studies.  
In addition, neurodegenerative diseases such as Alzheimer's disease can cause temporal 
shape changes as the disease progresses. 
Finally, image acquisition and processing algorithms for extracting shape information can cause measurement variation in the estimated shapes.

While variation can be incorporated into shape models via different approaches, e.g.
via random initial conditions as in the random orbit model \cite{miller_statistical_1997} 
or in a mixed-effects setting \cite{allassonniere_towards_2007}, most models
specify random variation relative to a base object, usually a template, and they 
involve some form of linearization. Here we will model these variations by inserting stochasticity directly into the nonlinear dynamics of shape transformations using Large Deformation Diffeomorphic Metric Mapping (LDDMM) \cite{younes_shapes_2010}. 

\begin{figure}[t]
	\centering
      \includegraphics[width=1\columnwidth]{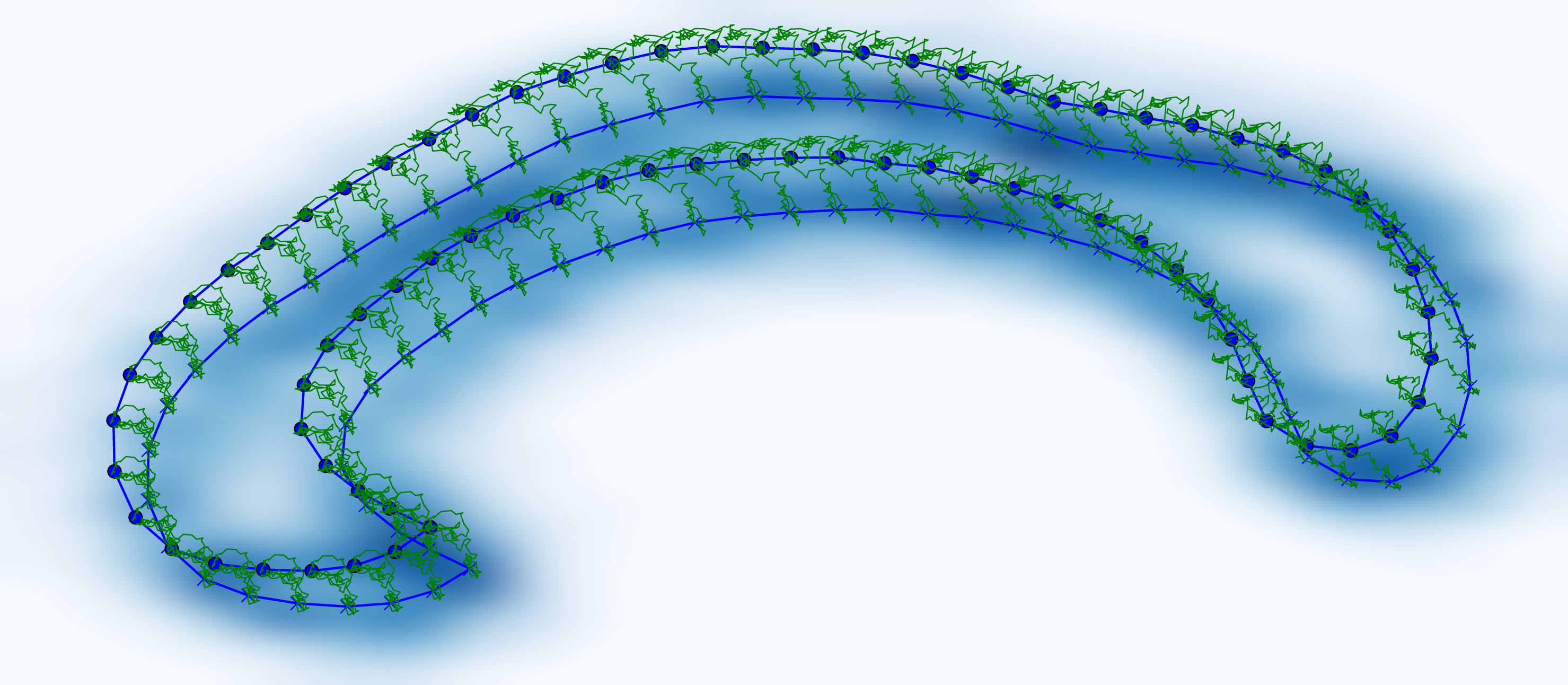}
  \caption{
    Sample from bridge process (green lines) conditioned on two corpus callosum
    shapes (blue). The landmarks are influenced by noise with spatial correlation
    structure determined by the noise fields $\sigma_l$. The bridge is
    constructed by integrating the perturbed gradient flow described in
    Section~\ref{sec:stochgradflow}. The shaded background indicates the
    shape variations in the transition distribution of the stochastic model.
    }
  \label{fig:intro}
\end{figure}

In this paper, we use the approach of \cite{arnaudon_stochastic_2016}, 
based on the stochastic fluid models of \cite{holm2015variational}, to
introduce a model for incorporating stochastic variation into the shape deformation paths.
 This approach is particularly designed to be compatible with four
geometric properties of LDDMM. Namely, (1) the noise is right-invariant in the same way as the
LDDMM metric is right-invariant. (2) Evolution equations arise as extremal paths for
a stochastic variational principle. (3) The Euler-Poincar\'e (EPDiff) 
equations in the deterministic case have stochastic versions. 
(4) Remarkably, the momentum maps arising via reduction by symmetry, and used to reduce dimensionality from the infinite-dimensional diffeomorphism group to finite-dimensional shape spaces, such as landmarks in deterministic LDDMM, still persist in this stochastic geometric setting.

{\bf Plan.} After a short description of large deformation models in Section~\ref{background} and the 
LDDMM framework, we discuss the model of \cite{arnaudon_stochastic_2016} in the CA 
context and use it to develop the stochastic equations for the landmark trajectories in Section~\ref{sto-landmark}.
We then formulate two approaches for estimating the noise fields and initial
shape configuration and momentum from prescribed data in Section~\ref{estimation}. 
The first approach is a deterministic
approximation of the Fokker-Planck equation that allows matching of moments
of the transition distribution. The second is an Energy-Maximization (EM) algorithm that
requires sampling of diffusion bridges. From the bridge simulation scheme, we
obtain (see Fig. \ref{fig:intro}) a stochastic version of the large deformation gradient flow algorithm, see
e.g. \cite{younes_shapes_2010}. In
particular, we can estimate data likelihood by sampling bridges without
registering or matching the landmarks.
Finally, we validate the approach in experiments on synthetic and real data in Section~\ref{example} before
providing outlook and concluding remarks in Section~\ref{conclusion}.

\section{Background}\label{background}

This section briefly reviews large deformation shape modelling and LDDMM, referring to the monograph  \cite{younes_shapes_2010} for full details.
A shape, defined as either a subset $s\subset\mathbb R^d$, $d=2,3$, or an image $I:\mathbb
R^d\rightarrow\mathbb R$ can be modified by warping the domain $\mathbb R^d$ 
under the action of a diffeomorphism $\varphi:\mathbb R^d\rightarrow\mathbb R^d$. The diffeomorphism 
$\varphi$ naturally acts on the left on a shape $s$ by $\varphi.s=\varphi(s)$. For images, a
left-action is obtained by letting $\varphi.I=I\circ\varphi^{-1}$. 
Changes in shape are then represented by an element  
$\varphi\in G_V\subset\mathrm{Diff}(\mathbb R^d)$ in the diffeomorphism group.
The subset $G_V$ is obtained as endpoints of flows
\begin{equation}
  \partial_t\varphi_t(x)=v_t\circ\varphi_t(x),\ t\in[0,T]
  \label{eq:lddmm-ode}
\end{equation}
where $v_t\in V$ is a time-dependent vector field contained in a space
$V\subset\mathfrak X(\mathbb R^d)$. $V$ is typically a reproducing kernel
Hilbert space (RKHS) with inner product $\left<\cdot,\cdot\right>_V$ defined in
terms of a reproducing kernel 
$K_V:\mathbb R^d\times\mathbb R^d\rightarrow\mathbb R^{d\times d}$ and with
corresponding momentum operator $L_V:V\rightarrow L^2$. Under reasonable
assumptions, the RKHS structure
defines a Riemannian metric on $G_V$ by right-invariance, i.e. by defining
$  \left<v_\varphi,w_\varphi\right>_\varphi
  =
  \left<v_\varphi\circ\varphi^{-1},w_\varphi\circ\varphi^{-1}\right>_V$
for tangent vectors $v_\varphi,w_\varphi\in T_\varphi G_V$. The corresponding Riemannian 
metric is
\begin{equation}
  d_{G_V}(\varphi,\psi)^2
  =
  \min_{v_t,\varphi_0=\varphi,\varphi_1=\psi}\int_0^T\|v_t\|_{\varphi_t}^2dt
  \label{eq:dgv}
\end{equation}
where $\varphi_t$ evolves according to \eqref{eq:lddmm-ode}. 
Elements of $G_V$ act on shapes, and $d_{G_V}$ descends to
shape spaces by
$  d_S(s_1,s_2)
  =
  \min_{\varphi.s_1=s_2}d_{G_V}(\mathrm{Id},\varphi)
    $
and similarly for images. Shapes can be matched by finding a minimising
$\varphi$ in this equation. In inexact matching of shapes, the distance is
weighted against a
dissimilarity measure. For images, the inexact matching problem becomes
  $\varphi_\mathrm{min}= \mathrm{argmin}\,E(I_1,I_2,\varphi)$,
  where 
  $E(I_1,I_2,\varphi)
  =
  d_{G_V}(\mathrm{Id},\varphi)^2
  +\lambda S(\varphi.I_1,I_2)$.
Here $\lambda>0$ controls the dissimilarity penalty, and $S$ is the 
dissimilarity measure, e.g.
the $L^2$ difference $S(I_1,I_2)=\int_{\RR^d}|I_1(x)-I_2(x)|^2dx$.
Optimal paths in the LDDMM framework satisfy the EPDiff equation $\partial_t m  = -\,\mathrm{ad}^*_{u} m$, where $m= L_V u$ and $\mathrm{ad}^*$ is the coadjoint action on the dual of the Lie algebra of vector fields on the plane, see \cite{HoMa2004} for more details.

For a set of $N$ landmarks $\mathbf q\in\mathbb R^{dN}$ at positions $q_1,\ldots,q_N\in\mathbb R^{d}$ with momentum $\mathbf p=(p_1,\cdots,p_N)\in \mathbb R^{dN}$, the singular momentum map of \cite{HoMa2004} is given by
$
m(x,t) = \sum_i p_i(t) \delta\big( \mathbf{x}-\mathbf{q}_i(t)\big)
$.
The corresponding deformation velocity can be written as $u(x,t)= K_V*m = \sum_i p_i K_V(x-q_i)$.
The ODE for the positions and momentum are the canonical Hamiltonian equations 
$\mathbf{\dot{q}}_i = \frac{\partial h}{\partial \mathbf p_i}$ and $ \mathbf{\dot{p}}_i = -\frac{\partial h}{\partial \mathbf q_i} $,
with Hamiltonian $h(p,q)= \frac12 \sum_{i,j=1}^N p_i\cdot p_j K_V(q_i-q_j)$.
The 1D versions of these equations go back to the dynamics of singular solutions of the Camassa-Holm equation \cite{camassa1993}.

\section{Stochastic landmark dynamics}\label{sto-landmark}

Stochastic differential equations (SDEs) have previously been considered in context of LDDMM 
by \cite{TrVi2012,vialard2013extension,cotter2013bayesian}. 
More recently \cite{marsland2016langevin} introduced a stochastic landmark model with dissipation in the momentum equation which corresponds to a Langevin equation. 
This allowed then to use the technology of statistical mechanics and Gibbs measure to study this stochastic system. 
We will use here the stochastic model of \cite{arnaudon_stochastic_2016}, based on the fluid models of \cite{holm2015variational}.
In short, the noise is introduced in the reconstruction relation \eqref{eq:lddmm-ode} in the following sense
\begin{equation}
    d \varphi_t\circ\varphi_t^{-1}(x)= v_t(x)dt +\sum_{l=1}^J \sigma_l(x)\circ dW_t^l\, ,
  \label{eq:lddmm-ode-sto}
\end{equation}
where we have introduced $J$ arbitrary fields $\sigma_l(x):\mathbb R^d \to V$. Here, the symbol $d$ denotes stochastic evolution and $(\,\circ\,)$, when adjacent to the process $dW_t^l$, means stochastic integrals are interpreted in the Stratonovich sense.  
Given a realisation of the noise, we impose the stochastic dynamics \eqref{eq:lddmm-ode-sto} and seek the $\varphi_t$ that minimises the cost functional in \eqref{eq:dgv}.
A short computation gives the stochastic EPDiff equation
    $dm + \mathrm{ad}^*_{v} m dt+ \sum_i \mathrm{ad}^*_{\sigma_i} m \circ dW^i_t=0  $.
As in the deterministic case, the landmarks can be introduced, and their
stochastic ODE dynamics are given by
\begin{align}
    \begin{split}
	    d \mathbf q_i &= \frac{\partial h}{\partial \mathbf p_i} dt + \sum_{l=1}^J\sigma_l(\mathbf q_i) \circ d W_t^l\\
	    d \mathbf p_i &= -\frac{\partial h}{\partial \mathbf q_i} dt - \sum_{l=1}^J \frac{\partial}{\partial \mathbf q_i}\left (\mathbf p_i \cdot\sigma_l(\mathbf q_i)\right )  \circ  d W_t^l
        \ .
    \end{split}
    \label{sto-Ham}
\end{align}
This is a Hamiltonian stochastic process in the sense of Bismut \cite{bismut1982mecanique}, where the stochastic extensions of the Hamiltonian, or stochastic potentials, are the momentum maps $\Phi_l(\mathbf q, \mathbf p) = \sum_i \mathbf p_i \cdot\sigma_l(\mathbf q_i)$ whose Hamilton equations generate the cotangent lift of the stochastic infinitesimal transformations of the landmark paths.

More details about the derivation of these equations can be found in \cite{arnaudon_stochastic_2016}. 
The noise can be seen as almost additive in the position equation, and it couples to the momentum with the gradient of the noise field $\sigma_l(\mathbf q_i)$ in the momentum equation.  
This last term ensures that the solution corresponds to a stochastic path in the diffeomorphism group. 
By comparison with previous works such as \cite{TrVi2012,vialard2013extension,marsland2016langevin}, this model has its noise amplitudes $\sigma_l$ fixed to the domain, and not to each landmark. The noise is thus relative to a Eulerian frame and is right-invariant similar to the metric.

\subsection{The Fokker-Planck equation}

For each time $t$, we will denote the transition probability density by $\mathbb P(\mathbf q,\mathbf p,t)$, a function $\mathbb P:\mathbb R^{2dN} \to \mathbb R$ which integrates to $1$ and represents the probability of finding the stochastic process at a given position in the phase space. 
For a given initial distribution $\mathbb P_0$, it is possible to compute the partial differential equation which governs the evolution of $\mathbb P$ in time. 
Using the canonical Hamiltonian bracket
$\{F,G\}_\mathrm{can} =\sum_i \frac{\partial F}{\partial q_i} \frac{\partial G}{\partial p_i} - \sum_i  \frac{\partial F}{\partial p_i} \frac{\partial G}{\partial q_i}$, 
we directly compute the Fokker-Planck equation in a compact form as  
\begin{align}
	\frac{d}{dt} \mathbb P = \{\mathbb P,h\}_\mathrm{can} + \frac12 \sum_{l=1}^J\{ \Phi_l, \{\Phi_l,\mathbb P\}_\mathrm{can}\}_\mathrm{can}\, . 
\end{align}
The right-hand side of this equation has two parts, the first is an advection with first order derivatives of $\mathbb P$, arising from the deterministic Hamiltonian dynamics of the landmarks. 
The second part is a diffusion term with second order derivatives of $\mathbb P$ which arises only from the noise.
This is the term which will describe the error in each landmark path subject to noise.

\subsection{Equivalent formulation with `Lagrangian noise'}

The covariance between the stochastic displacements
$d\mathbf q_i$, $d\mathbf q_j$ conditioned on the position of two landmarks 
$\mathbf q_i$, $\mathbf q_j$ may be computed as
\begin{align*}
  \mathrm{Cov}(d\mathbf q_i|\mathbf q_i,d\mathbf q_j|\mathbf q_j)
  =
  \sum_{l}
  \sigma_l(\mathbf q_i|\mathbf q_i) 
  \sigma_l(\mathbf q_j|\mathbf q_j) 
  \mathrm{Var}(dW_t^l
  )
  \ ,
\end{align*}
where $\mathrm{Var}(dW_t^l)$ is the variance of the $l$-th Brownian motion $dW_t^l$.
Thus, for a finite time discretization of the process in It\^o form and an increment $\Delta=[t_1,t_2]$,
$\mathrm{Cov}(\Delta\mathbf q_i|\mathbf q_i,\Delta\mathbf q_j|\mathbf q_j)=
  \sum_{l}
  \sigma_l(\mathbf q_i|\mathbf q_i) 
  \sigma_l(\mathbf q_j|\mathbf q_j) 
  (t_2-t_1)$. The stochastic differential $d\mathbf q$ can therefore formally be viewed
  as a Gaussian process on $\mathbb R^d$. The $dN\times dN$ matrix
    $\sigma^2(\mathbf q)
    =
      \sum_{l}
    \left[
      \sigma_l(\mathbf q_i) 
      \sigma_l(\mathbf q_j) 
    \right]^i_j$
  is symmetric. If $\sigma^2(\mathbf q)$ is positive definite,
  we let $K(\mathbf q)$ be a
  square root, i.e. $\sigma^2(\mathbf q)=K(\mathbf q)K(\mathbf q)^T$. 
  The following stochastic landmark dynamics is then equivalent to the original dynamics \eqref{sto-Ham}
\begin{align}
    \begin{split}
        d \mathbf q_i &= \frac{\partial h}{\partial \mathbf p_i} dt + \sum_{j,\alpha} K(\mathbf q)^i_{j ,\alpha}\circ d W_t^j\\
        d \mathbf p_i &= -\frac{\partial h}{\partial \mathbf q_i} dt - 
        \sum_{j,\alpha} \frac{\partial}{\partial \mathbf q_i}\left (\mathbf p_i \cdot 
        K(\mathbf q)^i_{j,\alpha}\right )  \circ  d W_t^{j,\alpha}
        \ ,
    \end{split}
    \label{sto-Ham-K}
\end{align}
where $j$ runs on the landmarks and $\alpha$ the spatial dimensions. 
Note that in \eqref{sto-Ham-K}, $W_t\in\mathbb R^{dN}$ as compared to $\mathbb
R^J$ previously. 
With this approach, instead of starting with the spatial noise fields 
$\sigma_1,\ldots,\sigma_J$, we can
specify the stochastic system directly in terms of the matrix $K(\mathbf q)$.
One natural choice is to set 
  $[K(\mathbf q)]^i_j
  =
  \mathrm{Id}_dk(\mathbf q_i-\mathbf q_j)
    $
for some scalar kernel $k$ and where $\mathrm{Id}_d$ is the $d\times d$ identity matrix. The possible reduction in dimensionality of the
noise from $J$ to $dN$ has computational benefits that we will exploit in the
following.

\section{Estimation of Noise and Initial Conditions}\label{estimation}
We now aim for estimating a set of parameters for the 
model from $N$ sets of observed
landmarks $\mathbf q_1,\ldots,\mathbf q_{N}$ at time $T$.
The parameters can be both parameters for the noise fields as described below
and the initial landmark configuration $\mathbf q_0$ and momentum $\mathbf p_0$.
The initial configuration $\mathbf q_0$ can be considered an estimated template 
of the dataset. The template will be optimal in the sense of reproducing the 
moments of the distribution
or in being fitted by maximum likelihood.

We use spatial noise fields $\sigma_1,\ldots,\sigma_J$ of the form
\begin{align}
    \sigma_l^\alpha (\mathbf q_i)
    = \lambda_l^\alpha  k(\|\mathbf q_i-\delta_l\|)\,,
    \label{kernel-gaussian}
\end{align}
where $\lambda_l\in \mathbb R^d$ is the spacial direction of the noise $\sigma_l$,
$\delta_l$ its centre, and the kernel
$k$ is either Gaussian 
$k(x)=e^{-x^2/(2r_l^2)}$ or cubic B-spline
$k(x)=S_3(x/r_l)$ with scale $r_l$.
The Gaussian kernels simplify
calculations of the moment equations. The B-spline representation has the
advantage of providing a partition of unity when used in a grid. The model is not limited to this set of noise functions.

\subsection{Matching of Moments}

Our first method relies on the Fokker-Planck equation of the landmarks and aims at minimising the cost functional
\begin{align}
    C(\braket{\mathbf p}(0), \lambda_l) = \frac{1}{\gamma_1} \left \|\braket{\mathbf q}_T- \braket{\mathbf q}(1)\right \|^2 + \frac{1}{\gamma_2}\left  \|\Delta_2\braket{\mathbf {qq}}_T- \Delta_2\braket{\mathbf {qq} }(1)\right \|^2,
    \label{cost-moment}
\end{align}
where $\braket{\mathbf p}(0)$ is the initial mean momentum, $\braket{\mathbf q}(1)$ the final mean position, $\braket{\mathbf q}_T$ the target sample mean position estimated from the observed landmarks, $\Delta_2 \braket{\mathbf {qq}}_T$ the centred sample covariance of the observations, $\Delta_2\braket{\mathbf {qq}}(1)$ the centered final covariance, and $\gamma_1,\gamma_2$ two parameters. 
The covariances are in components $\Delta_2\braket{q_i^\alpha q_j^\beta}$ and are $2\times 2$ matrices if $i=j$. 
For the norm, we used a normalised norm for each landmark such that they contribute equally in the sum. 
More explicitly, we have $\braket{q_i^\alpha }(t) := \int  q_i^\alpha \mathbb P(\mathbf q,\mathbf p,t) d\mathbf q d\mathbf p $, and $\Delta_2\braket{q_i^\alpha q_j^\beta }:= \braket{q_i^\alpha }\braket{ q_i^\beta }-\braket{ q_i^\alpha q_j^\beta }$.  
This cost functional corresponds to the error in matching the mean and covariance of the final probability density $\mathbb P(\mathbf q,\mathbf p,T)$.

It is not possible to solve the Fokker-Planck equation completely; so we will derive a set of ODEs that approximates the evolution of the mean and covariance by applying the standard cluster expansion method of quantum mechanics to the Fokker-Planck equation. 
We refer to \cite{arnaudon_stochastic_2016} for the explicit forms of these equations. 
The expected values of the higher-order products are approximated by only the $\Delta_2\braket{q_i^\alpha  q_j^\beta }$ and $\braket{q_i^\alpha}$ variables, upon neglecting higher-order correlations such as $\Delta_3\braket{q_i^\alpha q_j^\beta q_k^\beta }$.
In order to capture the effect of the noise in the momentum equation, the other correlations such as $\Delta_2\braket{ p_i^\alpha q_j^\beta }$ and $\Delta_2\braket{ p_i^\alpha p_j^\beta}$ must be taken into account. 
Their equations of motion are directly computed by taking the derivative in the definition of the correlation. 
For simplicity, we have used Gaussian kernels for both the Hamiltonian and noise fields so that derivatives have simple forms, and we have approximated $\braket{\sigma_l(\mathbf q)}\approx \sigma_l(\braket{\mathbf q})$.  
Higher-order corrections of this approximation are possible, but they would not result in significantly better results in practice. 

To avoid local minima in the minimization of \eqref{cost-moment}, 
we use a global optimisation method based on genetic algorithms, the differential evolution method \cite{storn1997differential}.
This method turns out to be rather successful for the examples presented below. 

\subsection{Maximum Likelihood}

We now derive a method for parameter estimation using maximum likelihood (ML).
Upon assuming the landmark observations are independent, the likelihood for the set of unknown 
parameters $\theta$ satisfies
$
  \mathcal{L}_k(\mathbf q_1,\ldots,\mathbf q_{N};\theta)
  =
  \prod_{i=1}^{N}
  \mathbb P(\mathbf q_i,T;\theta)
$
where $\mathbb P(\mathbf q,T;\theta)$ is the transition density at $\mathbf q$ marginalized over $\mathbf p$ and with parameters $\theta$. We will denote by $P(\mathbf q,\mathbf p;\theta)$ the law of the corresponding process.
An ML estimate of $\theta$ is then
$\hat{\theta}
  \in
  \mathrm{argmax}_\theta\,
  \mathcal{L}(\mathbf q_1,\ldots,\mathbf q_{N};\theta)
  $.
The classic EM algorithm \cite{dempster_maximum_1977} finds parameter estimates
$\theta_l$ converging to a $\hat{\theta}$ by iterating the following two steps: 
\begin{description}
  \item {\bf (E)} Compute expected log-likelihood 
    using parameter estimate $\theta_{l-1}$:
    \begin{equation}
        Q(\theta| \theta_{l-1})
        =
        \sum_{i=1}^{N}
        \mathbb  E_{P_{(\mathbf q,\mathbf p; \theta_{l-1})}}
        (
          \log \mathbb P(\mathbf q,\mathbf p; \theta|\mathbf q_i)
        )
        \,.
      \label{eq:Estep}
    \end{equation}
    \item {\bf (M)} Update the parameter estimate $\theta_l= \mathrm{argmax}_\theta\, Q(\theta|\theta_{l-1})$. 
\end{description}
We describe below a method for obtaining a Monte Carlo approximation of the conditional expectation given
$\mathbf q_i$ in \eqref{eq:Estep}.

\subsection{Stochastically Perturbed Gradient Flow}
\label{sec:stochgradflow}

The landmark trajectory $\mathbf q_i(t)$ for $t\in(0,T)$ and momentum $\mathbf p_i(t)$
for $t\in(0,T]$ can be considered the missing data for estimating the parameters $\theta$.
The conditional expectation in  \eqref{eq:Estep} amounts to marginalizing out the stochasticity when finding the expected log-likelihood of the full data. We approximate this marginalization by sampling diffusion
bridges, i.e. producing sample paths conditioned on hitting $\mathbf q_i$ at time T. In
\cite{delyon_simulation_2006,marchand_conditioning_2011}, a guidance scheme is constructed that modifies a
general diffusion process for a generic variable $\mathbf x$ conditioned on hitting a point $\mathbf v$ at time $T$ by adding a term of the form $\frac{\mathbf x-\mathbf v}{T-t} dt $ to the SDE.
This scheme allows sample based
approximation of $\mathbb E_\mathbf{x}(f(\mathbf{x})|\mathbf v)$ for functions $f$ of the original stochastic process $\mathbf x$ by the formula  
$\mathbb E_\mathbf {x}(f(\mathbf x)|\mathbf v)=C_\mathbf v\mathbb E_\mathbf{ y}(f(\mathbf y)\phi_\mathbf{ v}(\mathbf y))$ for the modified process $\mathbf y$ with a $\mathbf v$-dependent constant
$C_\mathbf v$ and a path-dependent correction factor $\phi_\mathbf{ v}(\mathbf y)$. 
We need to modify this scheme for application to landmarks, primarily because the diffusion field 
$\Sigma(\mathbf q,\mathbf p)$ in \eqref{sto-Ham} may not be invertible as required in the scheme of \cite{delyon_simulation_2006,marchand_conditioning_2011}. A scheme based on the pseudo-inverse $\Sigma^\dagger$ can be derived, but it is computationally infeasible for high-dimensional problems.
Instead, we construct the following landmark guidance scheme for the modified variables $\hat{\mathbf q}, \hat{\mathbf p}$
\begin{equation}
  \begin{pmatrix}
    d\hat{\mathbf q} \\
    d\hat{\mathbf p}
  \end{pmatrix}
  = \tilde{b}(\hat{\mathbf q},\hat{\mathbf p})dt
  -\frac{
      \Sigma^2 (\hat{\mathbf q},\hat{\mathbf p}) (\varphi_{T-t}(\hat{\mathbf q},\hat{\mathbf p})-\mathbf v
    )}{T-t}dt 
  + \Sigma(\hat{\mathbf q},\hat{\mathbf p})dW \, .
  \label{eq:boundedguid}
\end{equation}
Here, $\tilde{b}$ is a bounded approximation of the drift term in \eqref{sto-Ham} with the It\^o correction; $\varphi_{T-t}(\hat{\mathbf q},\hat{\mathbf p})$ is an approximation of the
landmark position at time $T$, given their position at time $t$; and $\Sigma^2 := \Sigma \Sigma^T$.
The boundedness condition together with appropriate conditions on $\Sigma$ is necessary to show that  $\lim_{t\rightarrow T}\hat{\mathbf q}=\mathbf v$ but it is not needed in practice. 
As in \cite{delyon_simulation_2006}, we can compute the correction factor $\phi_\mathbf{v}$
for the modified process \eqref{eq:boundedguid}. Because of the multiplication on $\Sigma^2$, the correction factor does not depend on the inverse or pseudo-inverse of $\Sigma$ and the scheme is thus computationally much more efficient.

Interestingly, the forcing term of \eqref{eq:boundedguid} 
is a time-rescaled gradient flow.
This can already be seen in the original scheme of \cite{delyon_simulation_2006}
by  noticing that 
$(\mathbf y-\mathbf v)=  \nabla_{\mathbf y} \frac12 \|\mathbf y-\mathbf v\|^2$ 
has the form of a gradient flow. 
In the present case, with appropriate conditions on
the noise fields $\Sigma(\mathbf q,\mathbf p)$, we can define
an admissible norm $\|\mathbf v\|_{\Sigma}^2:=\langle v, \Sigma^2 v \rangle $.
The forcing term in \eqref{eq:boundedguid} is then a derivative of a gradient flow for the norm 
$\|\varphi_{T-t}(\hat{\mathbf q},\hat{\mathbf p})-\mathbf v\|_{\Sigma}^2$ 
in the $\mathbf q$ variable. The gradient is taken with respect to the predicted 
endpoint $\varphi_{T-t}(\hat{\mathbf q},\hat{\mathbf p})$ and it couples to the $\mathbf p$ variable through $\Sigma^2$.
The flow is time rescaled with $t\to \frac12 (T-t)^2$, $dt\to (t-T)\,  dt$, $\Sigma\to \sqrt{T-t}\, \Sigma $ and $dW\to \sqrt{T-t}\, dW$.
This rescaling slows done the time when the original time $t$ is close to $T$.
This makes sure that the system has enough time to converge and reach the target $\mathbf v$. 
Deterministic gradient flows are used in greedy matching algorithms using
the LDDMM right-invariant metric as described in e.g. \cite{younes_shapes_2010}.
With the present stochastic flow,
sampling \eqref{eq:boundedguid} allows efficient evaluation of the $Q$-function in
EM and a similar evaluation of the data likelihood. In both cases, no matching or
registration of the data is needed.

\section{Numerical examples}\label{example}

In this section, we will illustrate the estimation algorithms on simple synthetic test
cases, and on a data set of corpora callosa shapes. 

\subsection{Synthetic Test Case}

This synthetic test case addresses matching between two ellipses discretized by 5 landmarks. 
We used a low number of landmarks here in order to have readable pictures.
(The algorithms scale well and are not limited by the number of landmarks.) 
For the noise fields $\sigma_l$, we use a grid of $4$ by $4$ Gaussian noise fields in the region $[-0.4,0.4]^2$ with a fixed scale $r_l = 0.085$ corresponding to the distance between the grid points. 
For testing purposes, we let the bottom $8$ of the noise fields have larger amplitude $\lambda_l$ in the $x$ direction, and the top $8$ fields have larger amplitudes in the $y$ direction.
With these parameters, we produced $5000$ sample landmark configurations from the model to estimate the sample mean and covariance of the landmarks at the final simulation time $T=1$.
\begin{figure}[t]
  \begin{center}
      \subfigure{\includegraphics[width=.45\columnwidth]{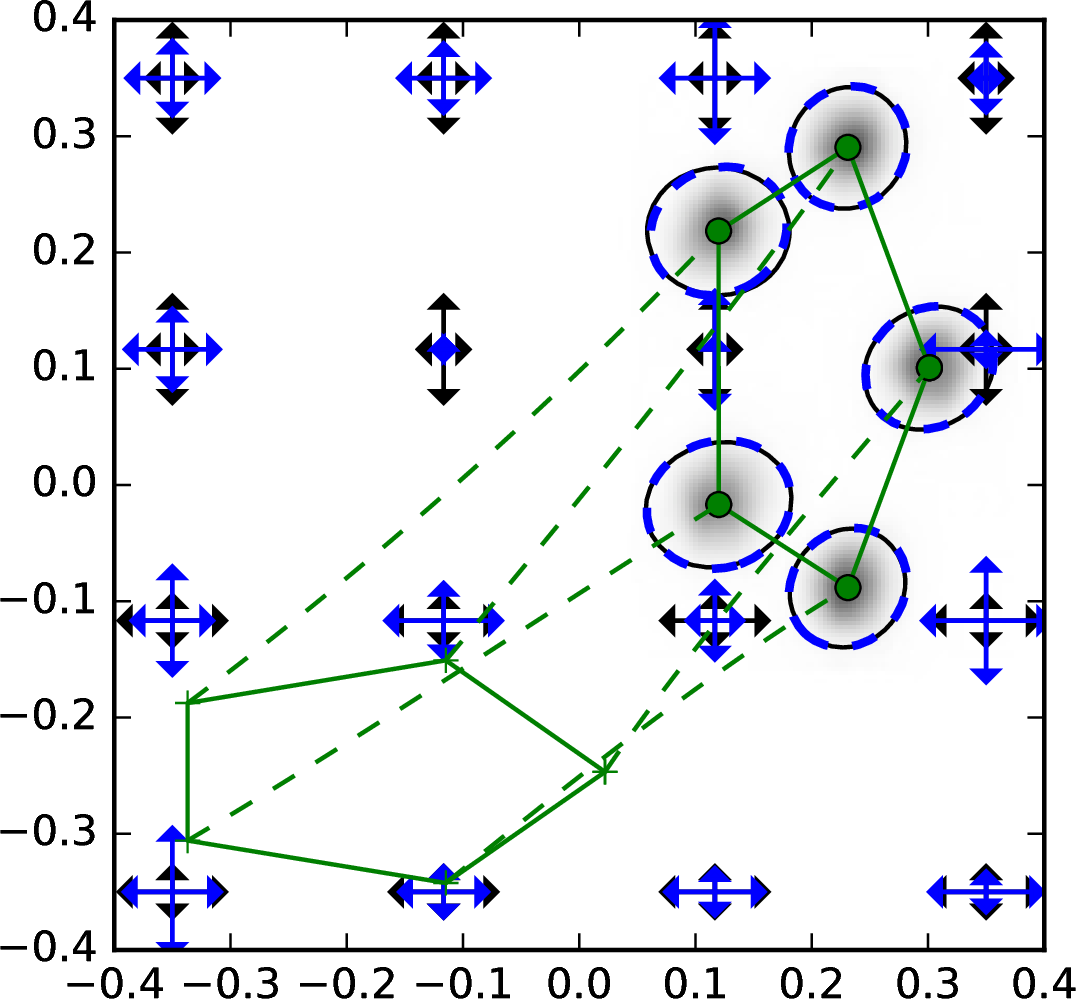}}
      \subfigure{\includegraphics[width=.40\columnwidth]{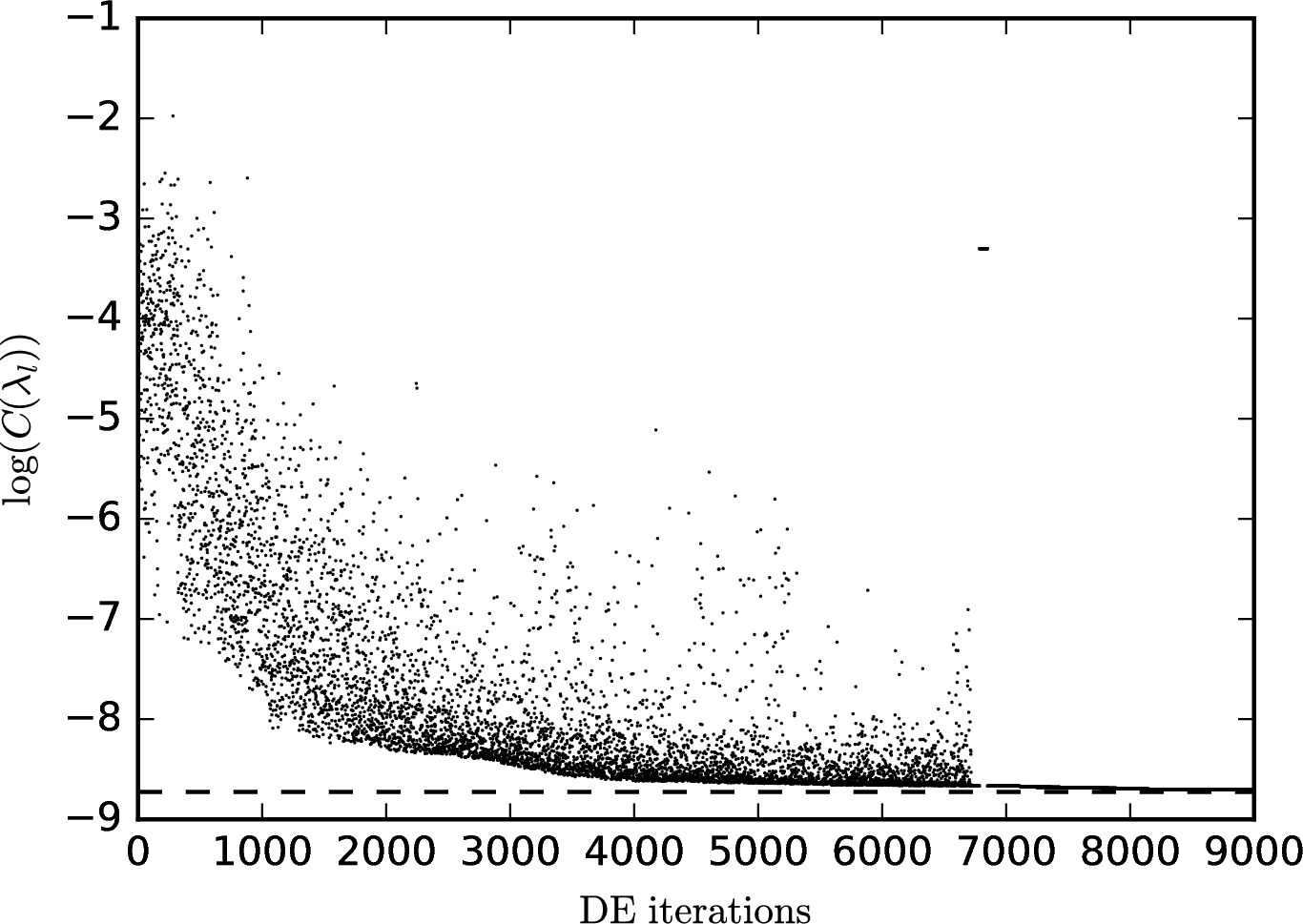}}
  \end{center}
    \caption{
        (left) Estimation of the noise amplitude (blue arrows) using only the final variance of the landmark (black ellipses).
        The landmarks trajectories (dashed green) have initial positions on the bottom left and the final mean position is indicated with green dots. 
        The amplitudes $\lambda_l$ of the noise fields are represented by arrows (black: original data, blue: estimated data). 
        We used the parameters $\alpha= 0.4$, $r_l= 0.085$, $\Delta t = 0.001$.
        (right) The convergence of the genetic algorithm which minimises the cost functional. For the last $2000$ points we used a gradient descent algorithm to improve the result.  
        } 
  \label{fig:ellipse-moment}
\end{figure}
On the left panel of Figure~\ref{fig:ellipse-moment}, we display the result of the moment algorithm. 
In black are shown the density and variance in black ellipses as well as the original fields $\sigma_l$ with the correct parameters.
The estimated parameters $\lambda_l$ of the noise fields $\sigma_l$ after running the differential evolution algorithm are shown in blue. The algorithm performs the estimation given only the final sample mean and covariance of the sampled landmarks.
The result is in good agreement for the final variances and shows small differences for the estimation of the parameters of the fields $\sigma_l$. 
The differences arise from three sources: the approximation used in the moment evolution; errors in the estimation of sample mean and variance; and the error in the solution of the minimisation algorithm. 
The minimisation algorithm may not have found the global minimum, but it did converge, as shown on the right panel of Figure~\ref{fig:ellipse-moment} where we display the value of the cost for each iteration of the genetic algorithm. 
Standard derivative-based algorithms would typically be caught in non-optimal local minima and thereby give worse results.

\subsection{Stochastic Gradient Flows}

\begin{figure}[t]
  \begin{center}
      \includegraphics[width=.40\columnwidth]{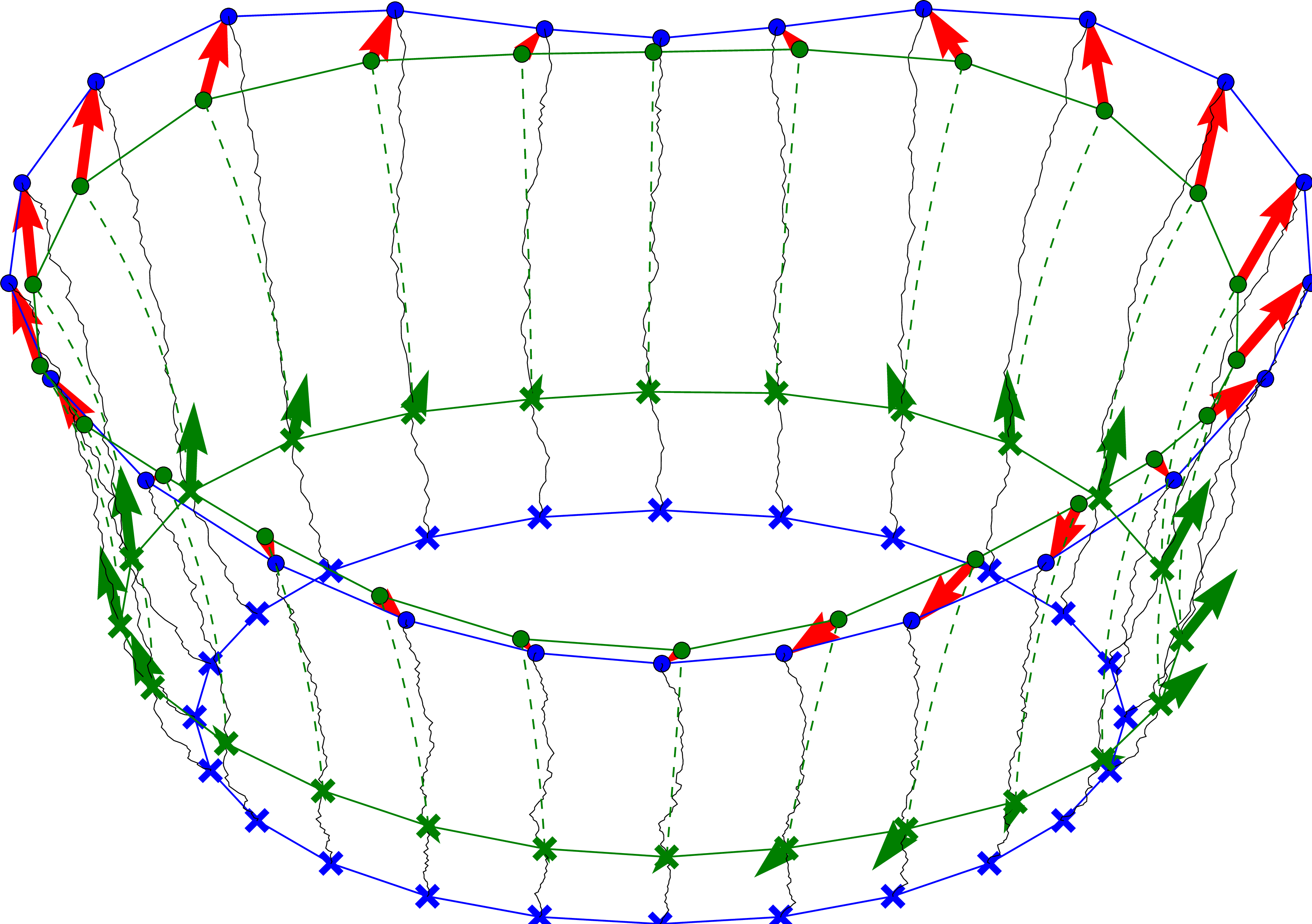}
      \qquad
      \includegraphics[width=.45\columnwidth]{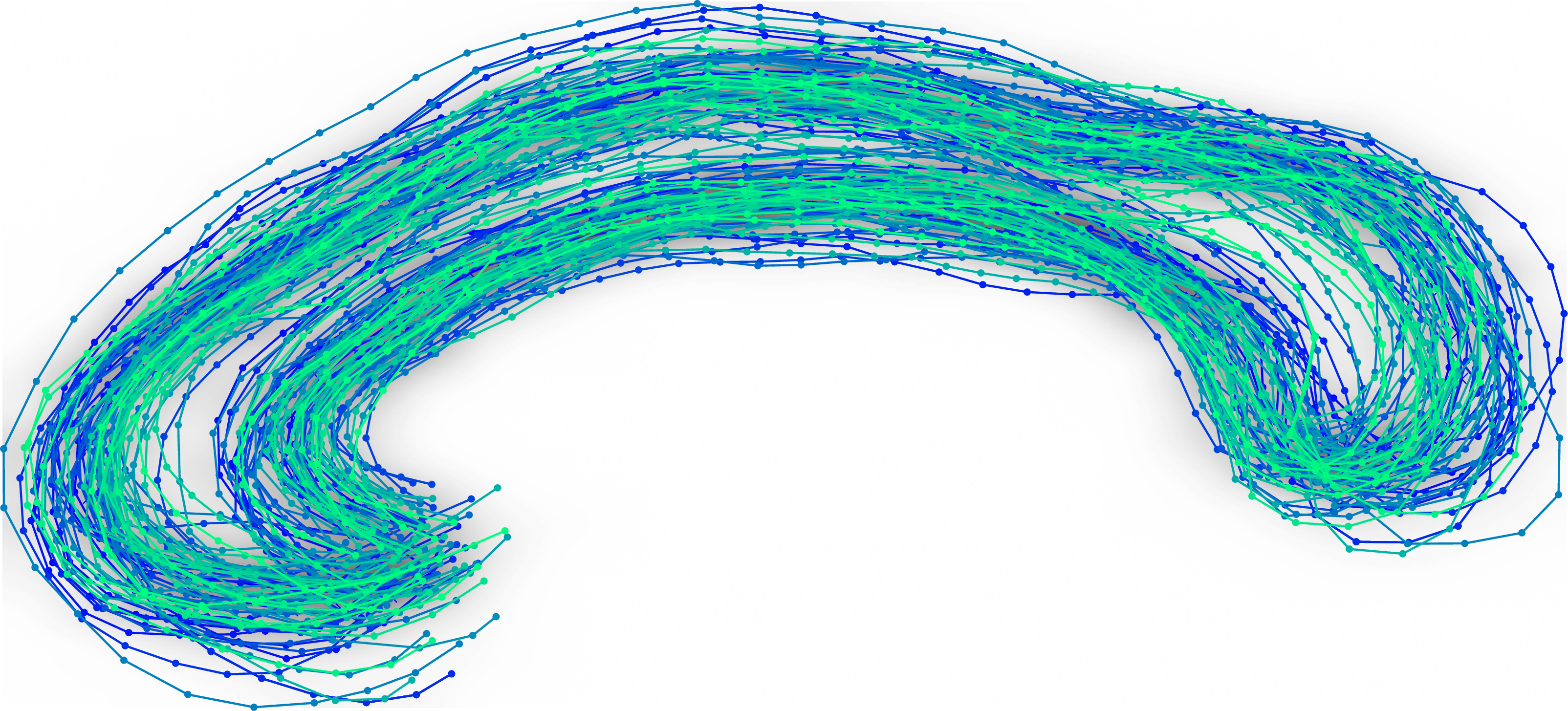}
  \end{center}
  \caption{
    (left) Stochastic gradient flow with guidance terms. At each time step $t$ with landmarks
    $\mathbf q(t)$ (green line+crosses), a prediction 
    $\varphi_{T-t}(\mathbf q(t),\mathbf p(t))$ of the endpoint configuration is computed 
    (green line+dots). From the prediction, the difference $\mathbf v-\mathbf q(t)$ to the target $\mathbf v$ (red arrows) is multiplied on $\Sigma^2$ and back transported 
    to $\mathbf q(t)$ giving the correction term (green arrows). 
    (right) The set of 65 corpora callosa shapes. Landmarks are evenly
    distributed along the outline of each shape.
    }
  \label{fig:sigmaxy}
\end{figure}

We consider matching two ellipses using the stochastically perturbed gradient
flow discussed in section \ref{sec:stochgradflow}. In Figure~\ref{fig:sigmaxy} (left), the initial set of landmarks 
$\mathbf q_0$ is displayed along the stochastic path
that is pulled by the gradient term to target set $\mathbf v$. 
The guidance is computed from the predicted endpoint 
$\varphi_{T-t}(\mathbf q(t),\mathbf p(t))$. The prediction and the guidance
term is showed at $t=T/4$. 
The guidance attraction in \eqref{eq:boundedguid} becomes increasingly strong with time as $t\to T$. This 
enforces $\mathbf q(t)\to\mathbf v$.

Compared to the scheme of \cite{delyon_simulation_2006}, the use of the function
$\varphi_{T-t}$ to predict the endpoint from $(\mathbf q(t),\mathbf p(t))$
removes much of the coupling between the momentum $\mathbf p$ and the guidance term. Without $\varphi_{T-t}$, the scheme would guide based solely on the difference $\mathbf v-\mathbf q(t)$. However, this would result in the scheme overshooting the target and producing samples of very low probability. The $\Sigma^2$ term on the guidance makes the scheme computationally feasible since the inverse of $\Sigma^2$ is not needed in the computation of the correction factor.

\subsection{Corpus Callosum Variation and Template Estimation}

\begin{figure}[t]
  \begin{center}
      \includegraphics[width=.59\columnwidth]{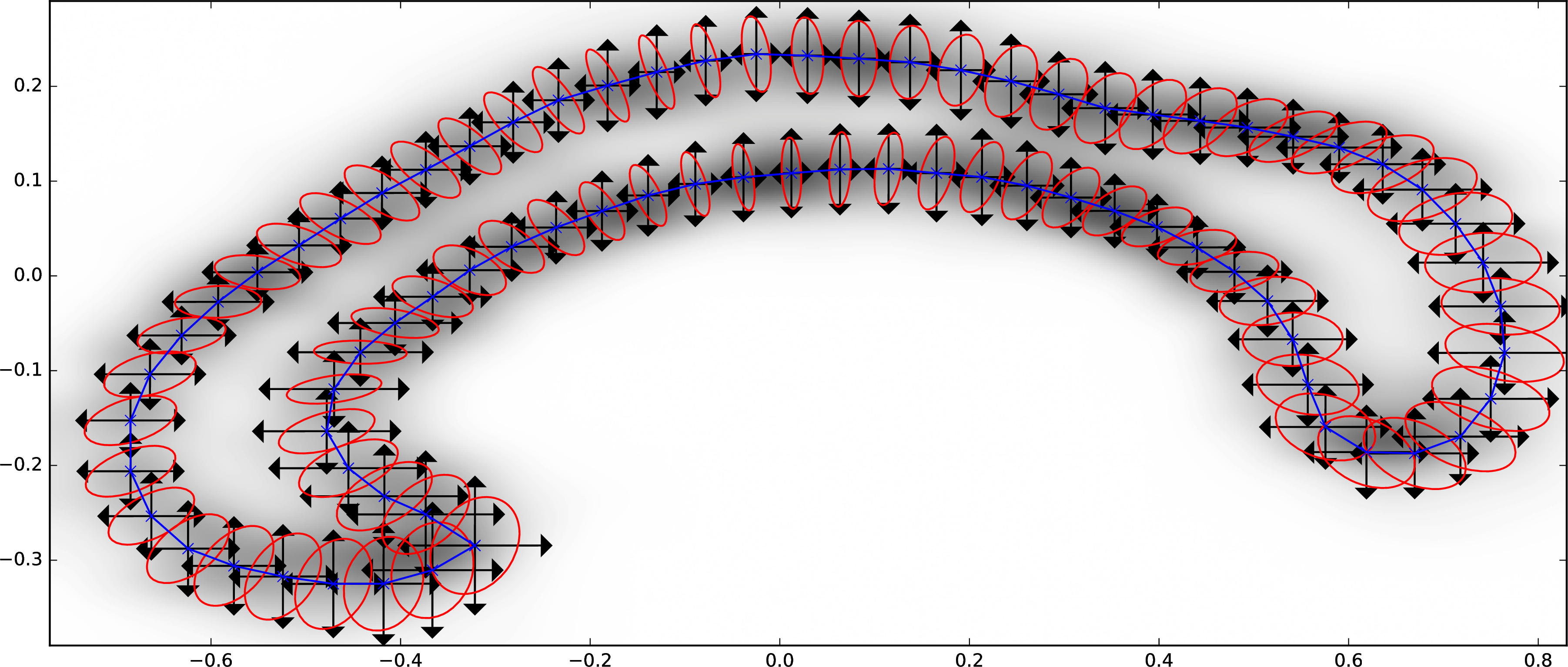}
	  {\includegraphics[width=.38\columnwidth]{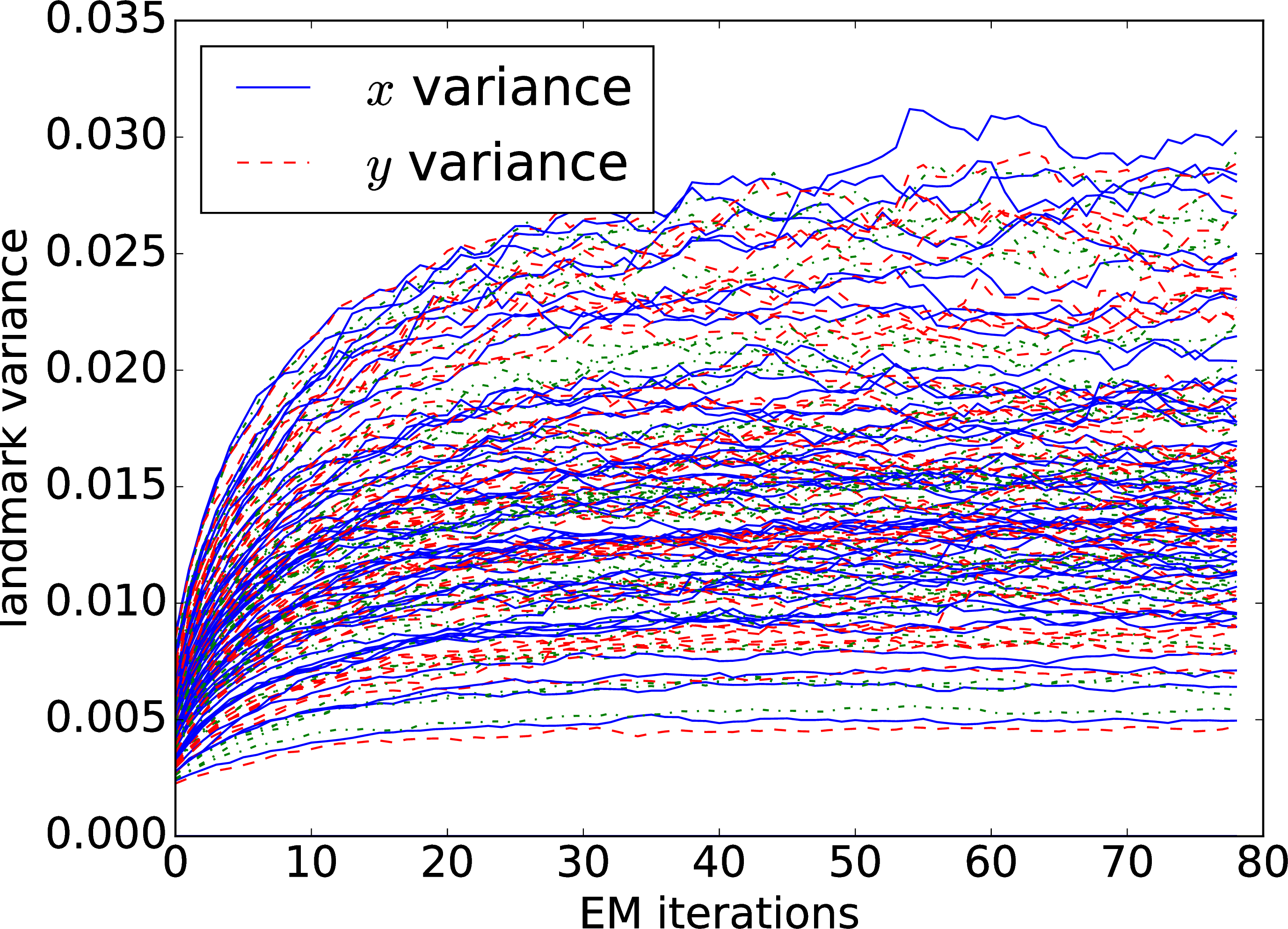}}
  \end{center}
  \caption{
    (left) Estimated template $\mathbf q_0$ from the corpus callosum shapes with variance magnitude given by arrow length at each landmark. 
    Sample noise covariances of each landmark of the original dataset plotted over each landmark (ellipsis). The 
    background is shaded with a smoothed histogram of the observed landmark positions.
    (right) Convergence of all variance parameters $(\lambda_l^0,\lambda_l^1)$ is shown as a
    function of EM iterations.
    }

  \label{fig:cc1}
\end{figure}
From a dataset of 65 corpus callosum shapes represented by 77 2D landmarks
$\mathbf q_{i,k}$, $i=1,\ldots,77$, $k=1,\ldots,65$
evenly distributed along the shape outline, here we aim at estimating the
template $\mathbf q_0$ and noise correlation represented in this  case
by a correlation matrix $K$. The parameters of the model are
$\theta=(\mathbf q_0,(\lambda_l^0,\lambda_l^1))$. For the MLE, we use the `Lagrangian' scheme 
with components of the spatial correlation matrix 
$k(\mathbf q_i,\mathbf q_j)=\mathrm{diag}(\lambda_j^0,\lambda_j^1)S_3(\|\mathbf
q_i-\mathbf q_j\|^2/r)$. The range
$r$ is $r=\mathrm{mean}_{ijk}\|\mathbf q_{i,k}-\mathbf q_{j,k}\|^2$. We initialize 
$\mathbf q_0$ with the Euclidean mean and run the EM algorithm for estimation of
$\theta$ until convergence. The evolution of the variances
$(\lambda_l^0,\lambda_l^1)$ is plotted in the right panel of Figure~\ref{fig:cc1}.
On the left panel of Figure~\ref{fig:cc1} shows the estimated template $\mathbf q_0$. The 
variance magnitude $(\lambda_l^0,\lambda_l^1)$
is plotted with arrows on each landmark and can be compared with the landmark-wise empirical variance 
from the dataset. The variance specified by $(\lambda_l^0,\lambda_l^1)$ is axis-aligned, and results in 
differences when the eigenvalues of the empirical variance are not axis-aligned.
\begin{figure}[t]
  \begin{center}
      \subfigure{\includegraphics[width=.65\columnwidth,trim=2 0 0 0,clip=true]{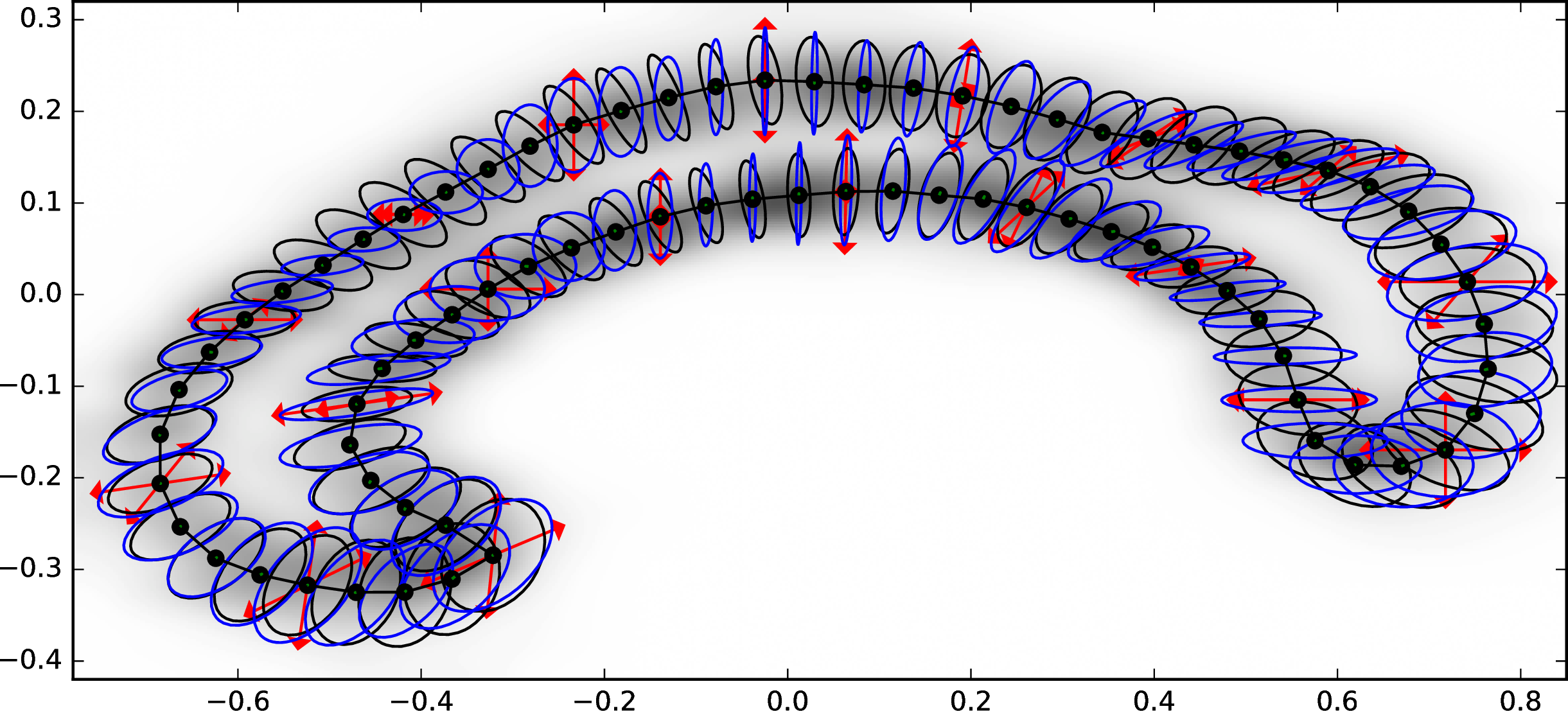}}
      \raisebox{-.2cm}{\subfigure{\includegraphics[width=.33\columnwidth,trim=0 0 0 0,clip=true]{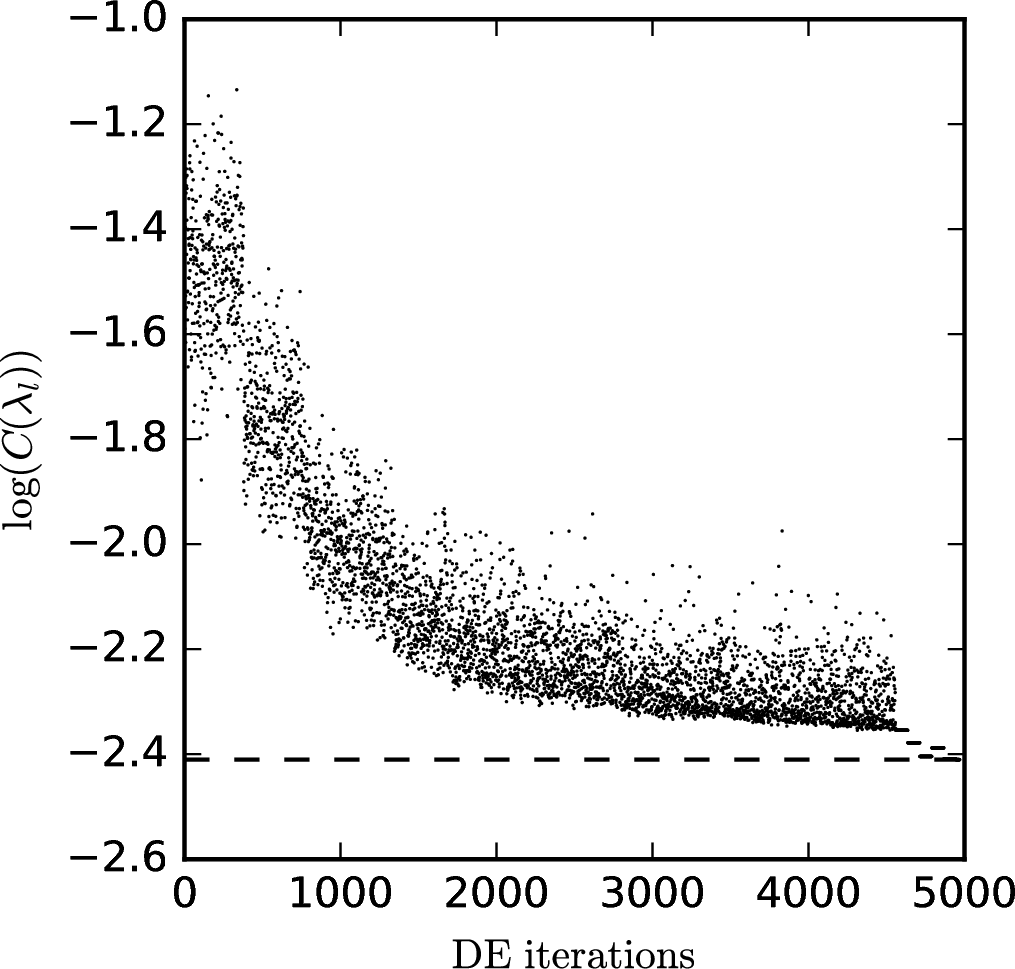}}}
  \end{center}
    \caption{Corpus callosum estimation with moment matching and setup as in Figure~\ref{fig:ellipse-moment}.
    We divided by $4$ the number of landmarks and placed the $\sigma$ at the same position. 
    We then ran the genetic algorithm to find the best set of $\lambda_l$ parameters which reproduced the variance for only the reduced set of landmarks. 
    From this set of $\sigma$, we used all the landmarks to compare the final estimated variance in blue with the observed variance in black. 
    The $\sigma$ fields are represented by red arrows. 
    We used the following parameters: $\alpha= 0.1$, $r_l= 0.1$, $\Delta t = 0.01$. 
  }
  \label{fig:cc-moment}
\end{figure}

This experiment is repeated in Figure~\ref{fig:cc-moment} with moment matching algorithm but with several differences of the model. 
First, the noise fields are the original spatially fixed Gaussian $\sigma_l$ represented by red arrows.
We also allow $\lambda_l$ to be non-axis aligned and we placed them at the position of every $4$ landmarks. 
We then applied the genetic algorithms using only the landmarks at the same position as the noise fields, fixing the initial momenta to $0$ and the initial position to the mean positions of the landmarks. 
In addition to its rapid convergence, as shown in the right panel of Figure~\ref{fig:cc-moment}, the global minimisation algorithm also gave a reliable estimate of the final variance for all the landmarks, even when fewer landmarks and noise fields were used.

\section{Conclusion and Open Problems}\label{conclusion}

We have presented and implemented a model for large deformation stochastic variations in computational
anatomy. The extremal paths are governed by stochastic EPDiff equations which arise from a right-invariant stochastic variational principle, and admit reduction from the diffeomorphism
group to lower dimensional shape spaces (landmarks). 

We have shown that accurate estimation of the noise fields of the model can be achieved either by approximating the Fokker-Planck equations with a finite set of moments, or by Monte Carlo sampling and EM-estimation. 
We have derived and implemented the methods in both cases. The second approach introduces
the concept of stochastically perturbed gradient flows for data likelihood evaluation.

It can be hypothesised that shape evolution of human organs under the influence
of diseases do not follow smooth geodesics as in conventional models used in
computational anatomy, but rather it exhibits stochastic variations in shape as the disease
progresses. The approaches presented enable modelling of such variations. We
expect to test this hypothesis on additional shape spaces and with multiple
time point longitudinal shape datasets in future work.

\subsubsection*{Acknowledgements}

We are grateful to M. Bruveris, M. Bauer, N. Ganaba C. Tronci and T. Tyranowski for helpful discussions of this material.
AA acknowledges partial support from an Imperial College London Roth Award.
AA and DH are partially supported by the European Research Council Advanced Grant 267382 FCCA held by DH. DH is also grateful for support from EPSRC Grant EP/N023781/1.
SS is partially supported by the CSGB Centre for Stochastic Geometry and Advanced
Bioimaging funded by a grant from the Villum foundation.

\bibliographystyle{alpha}
\bibliography{biblio}

\end{document}